\pdfoutput=1
\documentclass[11pt]{article}

\usepackage[final]{acl}
\usepackage{times}
\usepackage{latexsym}
\usepackage[T1]{fontenc}
\usepackage[utf8]{inputenc}
\usepackage{microtype}
\usepackage{inconsolata}
\usepackage{graphicx}

\usepackage{amsmath}
\usepackage{adjustbox}
\usepackage{booktabs}
\usepackage{arydshln}
\usepackage{url}
\usepackage{multirow}
\usepackage{algorithm}
\usepackage{algpseudocode}

\title{Adaptive LoRA Merge with Parameter Pruning \\for Low-Resource Generation}

\author{Ryota Miyano \\
  Grad. Sch. of Information Science and Tech. \\
  Osaka University\\
  Japan \\
  \texttt{miyano.ryota@ist.osaka-u.ac.jp}\And
  Yuki Arase \\
  School of Computing\\
  Institute of Science Tokyo \\
  Japan \\
  \texttt{arase@c.titech.ac.jp} \\
}

\begin{document}
\maketitle
\begin{abstract}
This study proposes a simple yet effective LoRA merge method to achieve LLM adaptation for low-resource language generation tasks. 
The LoRA merge technique, which integrates multiple LoRA modules trained on different tasks, has gained attention as an effective and efficient approach for adapting LLMs to target tasks. 
However, previous methods are limited in adaptability as they keep the LoRA parameters frozen. 
Additionally, the low-resource problem has been out of their scope. 
We propose a LoRA merge method that updates and prunes LoRA parameters through fine-tuning with minimal target task data, which allows finer-grained adjustments of LoRA parameters and enhancement of task adaptability.  
Extensive experiments have been conducted taking summarization as a benchmark task. 
Our datasets cover various domains and multiple languages of English and Japanese. 
The results confirm that the proposed method achieves significant and consistent improvements in task adaptability over the previous methods. 
\end{abstract}

\section{Introduction}
The rapid advancements in Large Language Models (LLMs) have significantly enhanced text generation capabilities and performance across tasks such as translation, summarization, question answering, and code generation \citep{zhao2024surveyllm,raiaan2024reviewllm,minaee2024llmsurvey,qin2024large}.
However, LLMs often struggle with low-resource tasks, including those involving languages with scarce linguistic resources, specialized programming languages, or tasks in medical and other specialized domains \citep{nasution2024lowresource,shen2024language,cassano2024programming,singhal2023clinical}.
This performance degradation arises from the insufficient adaptation of LLMs to target tasks, despite their general knowledge obtained during pretraining.
Fine-tuning is a common method to enhance task-specific performance \citep{minaee2024llmsurvey,han2024parameter}, but its effectiveness is often constrained by limited training data in low-resource problems \citep{khade2024challenges,yang2024fine,to2024deakinnlp}.

An alternative approach gaining attention is the integration of multiple models, particularly using LoRA modules \citep{hu2022lora,mao2025survey,huang2024lorahub}.
For instance, combining a model with general language capabilities and another specialized in a specific task can improve performance on target tasks.
Such LoRA merge technique linearly combines LoRA modules into a single model. 
Existing studies \citep{zhao2024loraretriever,huang2024lorahub,wu2024mixture,wang2024loraflow} typically keep module parameters fixed and only adjust their combination weights, which reduces training costs. 
However, we assume it limits adaptability to the target task.
Furthermore, low-resource tasks have been out of their scope. 

To effectively adapt LLMs on low-resource language generation tasks, we propose a novel LoRA merge method that further updates LoRA modules with minimal target task data while pruning ineffective parameters.
% Additionally, we extend the LoRA merge process based on insights from prior research on the roles of different layers within LLMs.
Previous studies have reported that each decoder layer in LLMs plays a different role in language generation \citep{wendler2024llamalang}.
Furthermore, analyses of LoRA modules trained on multiple tasks suggest that these modules learn task-specific representations that vary across layers \citep{wu2024mixture}.
These findings inspired us to hypothesize that LoRA parameters may require finer-grained adjustments at different layers to better adapt to a target task. 
% optimal task performance may require different LoRA modules to be applied at different layers. 
Based on this hypothesis, our method evaluates the importance of each LoRA parameter at each layer while pruning away ineffective ones and retraining them in order to enhance task adaptability.
% We remove modules with minimal contributions and prioritize training of more effective modules, thereby improving task adaptability and performance.

We conducted extensive experiments to evaluate and analyze the proposed method taking summarization as a benchmark task. 
% on multiple language generation tasks, including English and Japanese summarization tasks.
Our datasets cover various domains of news, scientific papers, and radiology reports in multiple languages of English and Japanese. 
The results confirm that updating LoRA modules during the merge process improves task adaptability. 
In addition, pruning ineffective parameters further enhances the performance. 

The primary contributions of this study are twofold. 
First, our simple LoRA merge technique achieves effective LLM adaptation to low-resource tasks across various domains and multiple languages with a minimum amount of target-task data. 
Second, we show that LoRA parameter pruning enhances the task adaptability of LLMs, which is a novel feature of the pruning technique that often degrades the performance in exchange for the reduction of active parameters.  
The codes are available at \url{https://github.com/mr0223/adaptive_lora_merge}.

%%%%%%%%%%%%%%%%%%%%%%%%%%%%%%%%%%%%%%%%%%%%%%%%%%%%%%%%%%%%
\section{Related Work}
This section discusses the previous LoRA merge techniques. 
In addition, we review studies on LLM layer analysis that inspired us to conduct parameter pruning during the LoRA merging process. 

\paragraph{LoRA Merge.}
Several studies have investigated methods for combining multiple LoRA modules to facilitate multi-task learning.
Early approaches employed static integration strategies, such as averaging module outputs or using fixed, manually designed weights \citep{sun2023controlipe,smith2023construct}.
While these methods are computationally efficient, they often lack flexibility and struggle to adapt to tasks that differ significantly from those seen during training.
LoRAHub \citep{huang2024lorahub} addresses this limitation by optimizing integration weights while keeping the original LoRA modules frozen.
Task-specific LoRA modules are pre-trained on approximately $200$ tasks, and gradient-free optimization is applied to tune the integration weights based on a small number of target task examples.
% This approach enhances task-specific performance without requiring extensive retraining.
This data-efficient approach allows low-resource task adaptation. 
However, because LoRAHub relies solely on adjusting integration weights and keeping the LoRA modules frozen, its capacity to handle tasks that are highly distinct from the pre-training tasks is limited.

The proposed method builds on these approaches by overcoming their limitations.
Instead of relying solely on weights to combine frozen pre-trained modules, we directly update LoRA modules through target-task training with pruning for finer-grained adjustments of LoRA parameters. 
% This allows for improved task adaptation, particularly in low-resource scenarios where task-specific learning is critical.

\begin{figure}[t]
    \centering
    \includegraphics[width=\linewidth]{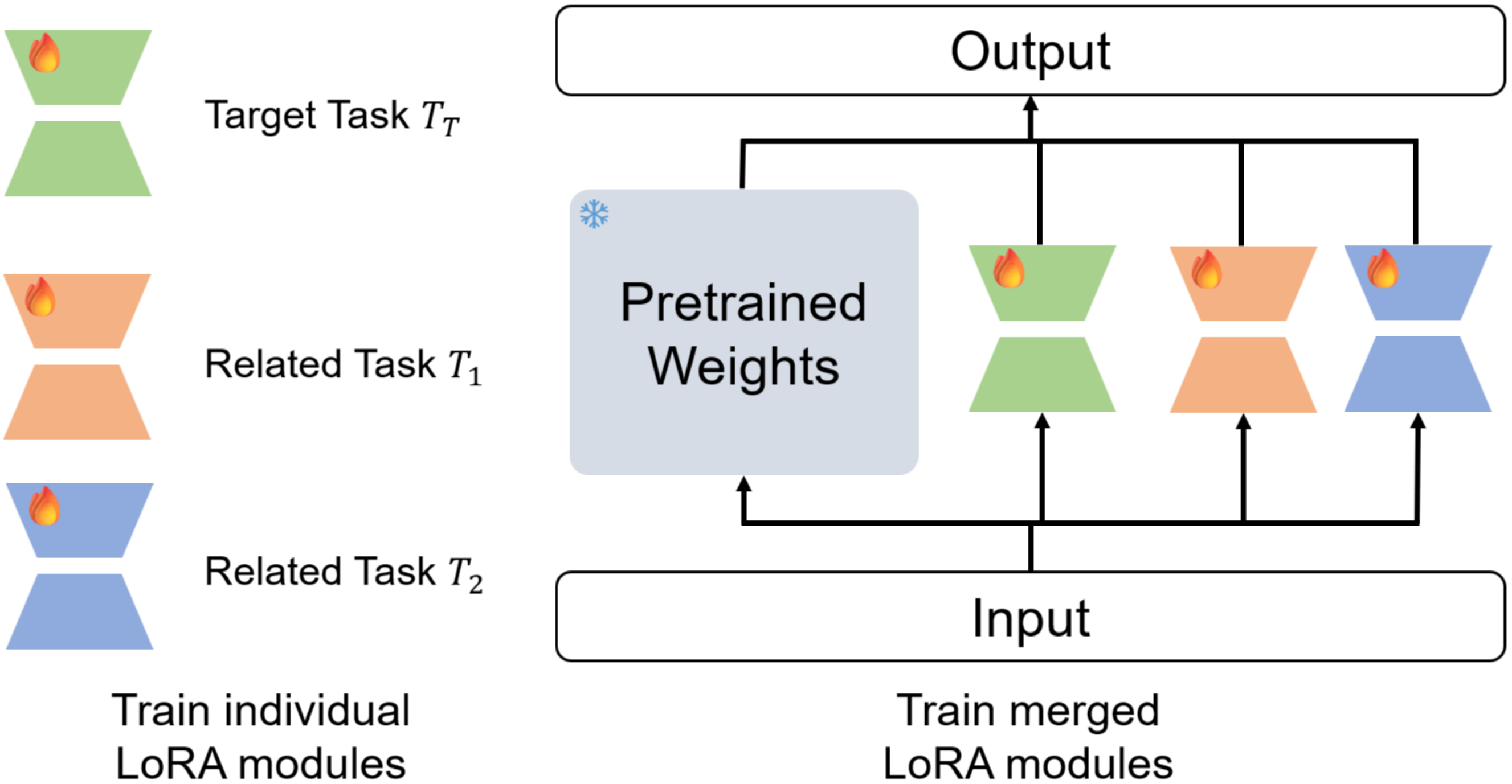}
    \caption{Two-stage training of LoRA modules: individual training on related tasks followed by fine-tuning with parameter pruning on a target task.}
    \label{fig:train_merge}
\end{figure}

\paragraph{LLM Layer Analysis.}
Emergent analyses of LLM layers have shown that different layers of language models play specialized roles in processing input data.
\citet{wendler2024llamalang} analyzed the Llama $2$ model \citep{meta2023llama2} and discussed that the layers conduct hierarchical processing to understand input texts.
% According to their findings, the input layers aggregate multiple input tokens to build feature representations, the middle layers transform these representations into abstract concept vectors suited to the task at hand, and the output layers convert these vectors into specific word sequences.
This hierarchical processing indicates that each layer contributes distinctively to tasks such as contextual understanding and language generation.
\citet{wu2024mixture} further investigated layer-specific characteristics in multi-task learning models utilizing LoRA modules.
They found that middle layers are more effective for simpler reasoning tasks, while upper layers are better suited to complex reasoning tasks.
Based on these observations, they proposed Mixture of LoRA Experts (MoLE) to improve the performance of multi-task learning. 
MoLE dynamically adjusts the integration of frozen LoRA modules by modifying module weights for each layer, and further, for each input text.
% MoLE enhances task-specific performance by leveraging the specialized roles of layers, yet the LoRA modules are kept frozen. 
% In addition, MoLE assumes the availability of abundant training data. 
MoLE enhances the multi-task learning performance; however, it assumes that abundant training data is available for the target task. 
% Our proposed method incorporates these insights by dynamically selecting and updating LoRA modules to align with the functions of individual layers.
% By tailoring module integration according to layer-specific characteristics, the method improves task adaptability and performance, particularly in scenarios with limited training data.
These studies inspired us to employ parameter pruning during LoRA merge to achieve finer-grained adjustments of LoRA modules for each LLM layer. 

%%%%%%%%%%%%%%%%%%%%%%%%%%%%%%%%%%%%%%%%%%%%%%%%%%%%%%%%%%%%
\section{Adaptive LoRA Merge with Pruning}
The proposed method achieves effective adaptation to a low-resource target task through training and pruning of LoRA parameters. 
Figure~\ref{fig:train_merge} illustrates the overview of the training procedure in the proposed method. 
The proposed method applies multiple LoRA modules trained on related tasks to a frozen LLM and further trains them on a target task (Section~\ref{subsection:lora_merge_training}).
During this process, the importance of LoRA parameters is evaluated at each decoder layer, and the parameters with lower importance are pruned and retrained (Section~\ref{subsection:delete_unnecessary_lora}). 
We remark that the proposed method does not explicitly `merge' LoRA parameters; rather, our merging process is implicit through updates and pruning of all the original LoRA parameters.

\subsection{Fine-Tuning of LoRA Modules}
\label{subsection:lora_merge_training}
First of all, individual LoRA modules are trained independently to learn related tasks on a frozen LLM. 
Then the proposed method adaptively merges these LoRA modules with further training. 

% LoRA is a parameter-efficient fine-tuning method that decomposes the weight update matrix $\Delta W$ into two low-rank matrices, $A$ and $B$, such that $\Delta W = BA$.
LoRA decomposes the weight update matrix of LLM, $\Delta W$, into two low-rank matrices, $A$ and $B$, such that $\Delta W = BA$.
We denote a LoRA module trained on a small set of target task data as $B^TA^T$, while we denote other $N$ LoRA modules trained on $N$ related tasks as $B_1A_1, B_2A_2, \dots, B_NA_N$. 
These modules are merged and then applied to the LLM parameters $W_0$, forming a new model parameterized as $W_0 + B^TA^T + B_1A_1 + \dots + B_NA_N$. 
This model is fine-tuned using the target task data, with the LLM parameters frozen.
The final parameters become $W_0 + \hat{B^T}\hat{A^T} + \hat{B}_1\hat{A}_1 + \dots + \hat{B}_N\hat{A}_N$, where $\hat{B^T}$, $\hat{A^T}$, $\hat{B}_i$ and $\hat{A}_i$ $(i = 1, 2, \dots, N)$ are the fine-tuned LoRA parameters on the target task. 

Note that the proposed method does not necessarily require $B^T$ and $A^T$. %, the LoRA module trained on the target task. 
It can instead rely on $N$ LoRA modules trained on other tasks. 
The effect of the target task LoRA is examined in our experiments. 

\subsection{Pruning of Ineffective LoRA Parameters}
\label{subsection:delete_unnecessary_lora}

\begin{figure}[t]
    \centering
    \includegraphics[width=\linewidth]{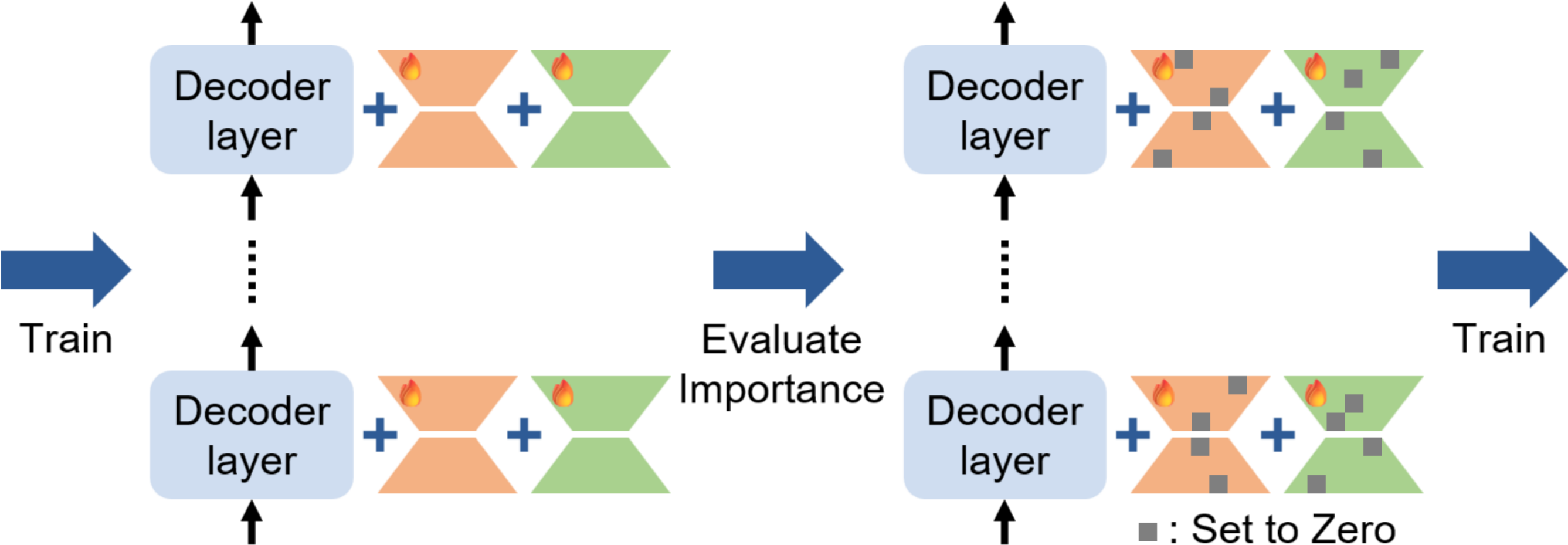}
    \caption{Pruning of LoRA parameters.}
    \label{fig:delete_module}
\end{figure}

Figure \ref{fig:delete_module} illustrates our pruning process.
During the training of merged LoRA modules, the importance of LoRA parameters is evaluated at each decoder layer, and ineffective parameters are pruned away at each training step. 
Algorithm~\ref{alg:pseudo_code} shows a pseudo-code of this process. 
After gradient calculation and parameter updates, parameters are evaluated for their importance. 
Ineffective parameters are pruned and then retrained at the next step. 

\begin{algorithm}
\caption{Adaptive LoRA merge with pruning}\label{alg:pseudo_code}
\renewcommand{\algorithmicrequire}{\textbf{Input:}}
\renewcommand{\algorithmicensure}{\textbf{Output:}}
\begin{algorithmic}
\Require Training and validation sets of target task $\mathcal{D}_t$ and $\mathcal{D}_v$, LMM $\mathcal{M}$ with frozen parameters $W_0$ and pre-trained LoRA modules $\mathcal{R}^{(0)}$
\Ensure LoRA modules with target task adaptation and pruning: $\hat{\mathcal{R}}^{(n)}$ 
\Repeat 
    \State Sample mini-batch $b_i$ from $\mathcal{D}_t$ for step $i$ 
    \State $\mathcal{L} \gets \mathcal{M}(b_i)$ \Comment{Compute loss}
    \State Compute gradients, backward loss $\mathcal{L}$ 
    \State $\mathcal{R}^{(i)} \gets \text{update} (\hat{\mathcal{R}}^{(i-1)})$ \Comment{Update LoRA}
    \State $\mathcal{E}^{(i)} \gets \text{eval}(\mathcal{R}^{(i)}, \mathcal{D}_v)$ \Comment{Eval. importance}
    \State $\hat{\mathcal{R}}^{(i)} \gets \text{prune}(\mathcal{R}^{(i)},\mathcal{E}^{(i)})$ \Comment{Pruning}
    \State $\mathcal{M} \gets W_0, \hat{\mathcal{R}}^{(i)}$ \Comment{Apply pruned LoRA}
\Until{converge}
\end{algorithmic}
\end{algorithm}

\begin{figure}[t]
    \centering
    \includegraphics[width=0.7\linewidth]{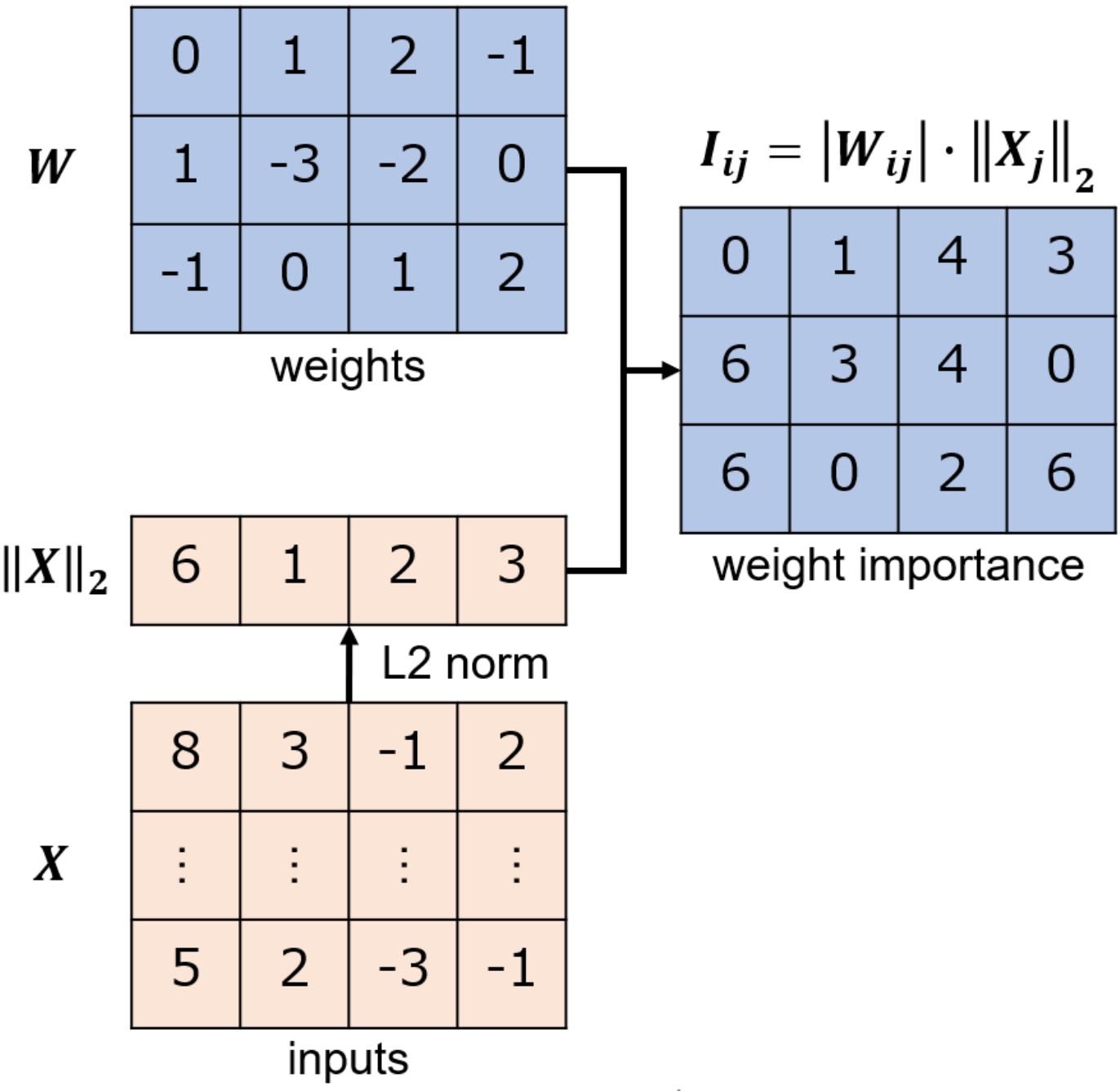}
    \caption{Importance calculation of LoRA parameters}
    \label{fig:calc_ipt_input}
\end{figure}

\paragraph{Parameter Importance}
Following \citep{sun2024wanda, dettmers2022llm}, we evaluate the importance of LoRA parameters based on the magnitude of parameter weights and inputs as illustrated in Figure~\ref{fig:calc_ipt_input}. 
\citet{sun2024wanda} empirically showed that not only the magnitude of parameters but also that of input activations should be considered because the scale of input features can significantly differ in LLMs. 
The importance is defined as the product of the absolute value of a parameter weight $W_{ij}$ and the $L_2$ norm of the corresponding input features:
\begin{align*}
    I(W_{ij}) = |W_{ij}| \cdot \|X_j\|_2
\end{align*}
where $|\cdot|$ computes the absolute value and $\|X_j\|_2$ is the $L_2$ norm of the associated input feature $X_j$.
The proposed method uses a validation set to compute the input features.

\paragraph{Pruning Strategy}
Low-importance parameters are pruned using a \textbf{zeroing} strategy; the weights of these parameters are set to zero and trained again in the next training step. %effectively removing their influence on the model.
This approach allows resetting parameters negatively affecting the target task performance and tuning them again, expecting they to learn better weights in the next step.

% \paragraph{Pruning Unit}
We conduct pruning at the parameter level, i.e., evaluating each parameter weight in a LoRA module individually and zeroing out low-importance ones. 
This approach is suitable when weight importance varies significantly within a LoRA module, as reported in  \citep{dettmers2022llm}. 
\citet{sun2024wanda} showed that parameter-wise pruning allows for retaining useful components while removing unnecessary sub-parameters. 
This can mitigate performance degradation due to excessive pruning by processing an entire module as a whole.

Weights are pruned based on a predefined ratio $s\%$; the lowest $s\%$ parameters in terms of importance are zero-out. 
As each LoRA module has been individually trained on different tasks, the distributions of parameter weights can vary across modules. 
Therefore, we compare the importance of parameters per each module rather than across modules. %, which has been confirmed as effective also in \citep{sun2024wanda}.
The pruning ratio is treated as a hyperparameter and optimized using validation data.

%%%%%%%%%%%%%%%%%%%%%%%%%%%%%%%%%%%%%%%%%%%%%%%%%%%%%%%%%%%%
\section{Experiment Settings}
We evaluate the capability of the proposed method for adapting an LLM for low-resourced target tasks. 
Intensive experiments are conducted using abstractive summarization as a benchmark task employing datasets of various domains of news, scientific papers, and radiology reports in multiple languages of English and Japanese.  

\subsection{Dataset}
This section provides an overview of the datasets used in our experiments, categorized into target and related tasks. 
The experiments cover both English and Japanese tasks.
The English tasks are summarization of radiology reports and scientific papers. 
The Japanese target tasks are summarization of research papers and news articles. 
Table~\ref{table:dataset} lists the number of data samples for each dataset.
Details on the construction and preprocessing of the target task datasets are provided in Appendix~\ref{appendix:dataset_construction}.

\subsubsection{Related Tasks}
We employed publicly available multilingual summarization datasets for pretraining LoRA modules of related tasks. 

\paragraph{XLSum}
The XLSum dataset \citep{hasan2021xlsum} is a multilingual news summarization dataset constructed from BBC news articles.
Both the Japanese and English subsets are used in our experiments.
Summaries are extracted from the lead sentences of the articles, which concisely present the main content of reported news.

\paragraph{WikiLingua}
The WikiLingua dataset \citep{ladhak2020wikilingua} is a multilingual resource derived from WikiHow guides.
Input documents consist of concatenated step explanations, while output summaries are formed by combining step headings.
We use both the Japanese and English subsets.

\subsubsection{Target Tasks}
For English tasks, we used two publicly available datasets distinct from the XLSum and WikiLingua domains. 
For Japanese, there is no available dataset for summarization other than XLSum and WikiLingua. 
Therefore, we created datasets for our experiments. 

\paragraph{MIMIC-III}
The MIMIC-III dataset \citep{johnson2016mimic} is used for the English radiology report summarization task.
Each report consists of three main sections: background, findings, and impressions.
The findings section serves as the input, and the impressions section, summarizing key observations, serves as the output.

\paragraph{SciTLDR}
The SciTLDR dataset \citep{cachola2020tldr} is used for the English scientific paper summarization task.
It contains short summaries (TLDRs) created by authors and reviewers.
The input consists of the abstract, introduction, and conclusion (AIC) sections, enabling the generation of highly compressed summaries.

\begin{table}[t]
    \centering
    \begin{adjustbox}{max width=\linewidth}
        \begin{tabular}{lrrr}
            \toprule
                Dataset & Train & Val & Test \\
            \midrule
                \multicolumn{4}{c}{Related task} \\
                XLSum (en) & $306,522$ & $11,535$ & $11,535$  \\
                XLSum (ja) & $7,113$ & $889$ & $889$  \\
                WikiLingua (en) & $98,999$ & $13,819$ & $28,607$  \\
                WikiLingua (ja) & $8,852$ & $1,264$ & $2,529$  \\
                \hdashline
                \multicolumn{4}{c}{Target task} \\
                MIMIC-III (en) & $44,342$ & $5,550$ & $10,996$ \\
                SciTLDR (en) & $1,992$ & $619$ & $618$ \\
                Bloomberg (ja) & $9,656$ & $1,207$ & $1,207$  \\
                NLP Paper (ja) & $312$ & $100$ & $100$  \\
                Medical Paper (ja) & $183$ & $100$ & $100$  \\
            \bottomrule
        \end{tabular}
    \end{adjustbox}
    \caption{Number of sentences in datasets}
    \label{table:dataset}
\end{table}

\paragraph{Bloomberg}
% This Bloomberg dataset was developed specifically for this study to support the Japanese news summarization task.
We crawled Bloomberg Japanese articles using the URL list provided by the MassiveSumm project \cite{varab2021massivesumm}. 
Bloomberg articles have bullet-point highlights that summarize the contents. 
We extracted them as ground-truth summaries combined with article titles.
The full article serves as the input document to summarize.
Remarkably, our way of dataset construction is different from that of XLSum utilizing lead sentences as summaries, to ensure that all the content in a summary exists in the input document. 
This difference makes Bloomberg task as distinct from XLSum, although the domain is the same. 
% Our approach can ensure that content written in summary exists in the input document. 

\paragraph{NLP/Medical Paper}
Two datasets were created from research papers on natural language processing and medical case reports. 
The task is generating titles from the corresponding abstracts as short summaries. 
The NLP paper dataset was built from the LaTeX corpus of the Journal of Natural Language Processing\footnote{\url{https://www.anlp.jp/resource/journal_latex/}}, extracting titles and abstracts.
The medical paper dataset was constructed from case reports published on J-STAGE\footnote{\url{https://www.jstage.jst.go.jp/}}, covering articles with diverse abstract formats.

\subsubsection{Evaluation Metrics}
The Bloomberg, MIMIC-III, and SciTLDR tasks were evaluated using ROUGE \cite{lin2004rouge}\footnote{\url{https://github.com/google-research/google-research/tree/master/rouge}}, while the NLP/Medical paper tasks were evaluated using BLEU \cite{papineni2002bleu}\footnote{\url{https://github.com/mjpost/sacrebleu}} due to their shorter summaries.  
For Japanese tasks, we employed the Mecab \citep{kudo-etal-2004-applying} for word segmentation. 
Additionally, statistical significance was assessed using approximate randomization testing \cite{riezler2005randomization}.

\subsection{Baselines}
\label{implementation}
We used the following baselines for comparison:
\begin{enumerate}
    \item \textbf{Zero-shot}: Summarization using an LLM without additional training.
    \item \textbf{LoRA (XS) / LoRA (WL)}: Summarization directly using LoRA modules trained on the related tasks of XLSum and WikiLingua, respectively.
    \item \textbf{LoRA (TGT)}: Summarization directly using LoRA modules trained on the target tasks.
\end{enumerate}

Additionally, we compare to LoRAHub, a strong baseline for LoRA merging.
LoRAHub involves merging LoRA modules from related tasks (denoted as ``\textbf{LoRAHub (XS+WL)}'') and further merging with the target task module (denoted as ``\textbf{LoRAHub (XS+WL+TGT)}''). %using $50$ samples of the target task training set. 
We reproduced LoRAHub based on its official Codes\footnote{\url{https://github.com/sail-sg/lorahub}}, making modifications to support Llama-$3$.

\begin{table*}[t]
    \centering
    \begin{adjustbox}{max width=\linewidth}
        \begin{tabular}{lcccccccccc}
            \toprule
                & \multicolumn{2}{c}{\bf MIMIC-III} & \multicolumn{2}{c}{\bf SciTLDR} & \multicolumn{2}{c}{\bf Bloomberg} & \multicolumn{2}{c}{\bf NLP Paper} & \multicolumn{2}{c}{\bf Medical Paper} \\
                \cmidrule(lr){2-3} \cmidrule(lr){4-5} \cmidrule(lr){6-7} \cmidrule(lr){8-9} \cmidrule(lr){10-11}
                & RL & Del\% & RL & Del\% & RL & Del\% & BLEU & Del\% & BLEU & Del\% \\
            \midrule
                Zero-shot     & $16.64$ & - & $29.58$ & - & $0.91$ & - & $2.73$ & - & $5.26$ & - \\
                LoRA (XS)     & $18.95$ & - & $24.76$ & - & $21.39$ & - & $12.26$ & - & $16.92$ & - \\
                LoRA (WL)     & $16.23$ & - & $33.23$ & - & $26.77$ & - & $18.89$ & - & $23.71$ & - \\
                LoRA (TGT)    & $27.97$ & - & $35.02$ & - & $25.64$ & - & $21.09$ & - & $30.95$ & - \\
            \hdashline
                LoRAHub (XS+WL)   & $18.83$ & - & $33.92$ & - & $27.11$ & - & $18.54$ & - & $23.66$ & - \\
                LoRAHub (XS+WL+TGT)    & $27.90$ & - & $\bf{35.63}^\dagger$ & -& $28.13^\dagger$ & - & $21.00$ & - & $26.93$ & -  \\
                Ours$_{\text{ Merge}}$ (XS+WL)      & $\bf{28.92}^\dagger$ & - & $\bf{35.95}^\dagger$ & - & $31.94^\dagger$ & - & $22.37^\dagger$ & - & $32.36^\dagger$ & - \\
                Ours$_{\text{ Merge}}$ (XS+WL+TGT)  & $\bf{29.13}^\dagger$ & - & $35.43$ & - & $31.79^\dagger$ & - & $22.46^\dagger$ & - & $30.86$ & - \\
                Ours$_{\text{ Merge+Del}}$ (XS+WL)  & $\bf{28.75}^\dagger$ & $30$ & $\bf{35.91}^\dagger$ & $30$ & $\bf{32.91}^\dagger$ & $40$ & $\bf{23.28}^\dagger$ & $50$ & $32.57^\dagger$ & $20$ \\
                Ours$_{\text{ Merge+Del}}$ (XS+WL+TGT)  & $\bf{28.96}^\dagger$ & $60$ & $\bf{35.99}^\dagger$ & $60$ & $\bf{33.12}^\dagger$ & $30$ & $\bf{23.04}^\dagger$ & $30$ & $\bf{34.04}^\dagger$ & $30$ \\
            \bottomrule
        \end{tabular}
    \end{adjustbox}
    \caption{Results on five summarization tasks of various domains and multiple languages. The best scores (scores with no significant difference from the highest ones) are marked by bold fonts, and $^\dagger$ indicates a significant difference against LoRA (TGT).}
    \label{table:main_result}
\end{table*}

\subsection{Implementation}
% The proposed method merges LoRA modules while incorporating a process to remove redundant modules during training.
% To evaluate its impact, we defined two configurations:
We evaluate variations of the proposed method to investigate the effects of LoRA fine-tuning on target tasks and parameter pruning of the proposed method: 
\begin{enumerate}
    \item \textbf{Ours$_{\text{ Merge}}$}: Conducts only fine-tuning of LoRA modules on target tasks.
    \item \textbf{Ours$_{\text{ Merge+Del}}$}: Conducts both LoRA fine-tuning and parameter pruning.
\end{enumerate}
% follows the same method used for baseline LoRA module training.
% In the Ours$_{\text{ Merge+Del}}$ setting, the deletion ratio was treated as a hyperparameter and optimized based on validation performance.
In Ours$_{\text{ Merge+Del}}$, the deletion ratio was treated as a hyperparameter and optimized based on the evaluation metrics measured on the validation data using grid-search.

For all the methods compared, we employed Llama-$3$-$8$B-Instruct~\citep{llama2024llama3}\footnote{\url{https://huggingface.co/meta-llama/Meta-Llama-3-8B-Instruct}} as the base model for its strong performance on various language tasks. 
The same prompt design was used for both LoRA module training and output generation.
We designed simple yet effective prompts tailored to each task to enhance learning and improve output quality.
The prompt details are provided in Appendix~\ref{appendix:implementation_details}.

\begin{table*}[t!]
    \centering
    \begin{adjustbox}{max width=\linewidth}
        \begin{tabular}{lp{0.7\linewidth}}
            \toprule
                Abstract    & Good representations facilitate transfer learning and few-shot learning. \textit{Motivated by theories of language and communication that explain why communities with large number of speakers have}, on average, simpler languages with more regularity, [...] Generalizing from there, we introduce \textbf{community-based autoencoders} in which \textbf{multiple encoders and decoders collectively learn representations} by being randomly paired up on successive training iterations. Our experiments show that [...] \\
                % Good representations facilitate transfer learning and few-shot learning. Motivated by theories of language and communication that explain why communities with large number of speakers have, on average, simpler languages with more regularity, we cast the representation learning problem in terms of learning to communicate. Our starting  point sees traditional autoencoders as  a single encoder with a fixed decoder partner that must learn to communicate. Generalizing from there, we introduce community-based autoencoders in which multiple encoders and decoders collectively learn representations by being randomly paired up on successive training iterations. Our experiments show that increasing community sizes reduce idiosyncrasies in the learned codes, resulting in more invariant representations with increased reusability and structure. \\
            \hdashline
                Reference   & \textit{Motivated by theories of language and communication}, we introduce \textbf{community-based autoencoders}, in which \textbf{multiple encoders and decoders collectively learn structured and reusable representations}. \\
                % Reference $2$       & The paper presents a \textbf{community based autoencoder} framework to address co-adaptation of encoders and decoders and aims at constructing better representations. \\
            \midrule
                Ours$_{\text{ Merge+Del}}$ (XS+WL+TGT)     & We introduce \textbf{community-based autoencoders}, a framework in which \textbf{multiple encoders and decoders collectively learn representations} by being randomly paired up on successive training iterations. \\
                Ours$_{\text{ Merge}}$ (XS+WL+TGT)     & \textbf{Community-based autoencoders} \textbf{learn more reusable and structured representations}. \\
                LoRAHub (XS+WL+TGT)    & We introduce a new framework for learning representations that is \textit{inspired by the way humans learn to communicate.} \\
                LoRA (TGT)    & We introduce a new framework for learning representations that is \textit{inspired by the way humans communicate} and learn from each other. \\
            \bottomrule
        \end{tabular}
    \end{adjustbox}
    \caption{Case study of the predicted output of different models (SciTLDR).}
    \label{table:example_output}
% \end{table*}
% \begin{table*}[t]
    \bigskip
    \centering
    \begin{adjustbox}{max width=\linewidth}
        \begin{tabular}{lllccccccccc}
            \toprule
                & & & \multicolumn{3}{c}{\bf Bloomberg} & \multicolumn{3}{c}{\bf NLP Paper} & \multicolumn{3}{c}{\bf Medical Paper} \\
                \cmidrule(lr){4-6} \cmidrule(lr){7-9} \cmidrule(lr){10-12}
                & & & RL & Thresh & Del\% & BLEU & Thresh & Del\% & BLEU & Thresh & Del\% \\
            \midrule
                \multicolumn{3}{l}{Ours$_{\text{ Merge}}$ (XS+WL+TGT)} & $31.79$ & -- & -- & $22.46$ & -- & -- & $30.86$ & -- & -- \\
            \hdashline
                \multirow{2}{*}{Input} & Zero & \multirow{4}{*}{Module}  & $32.01$ & $10$e-$3$ & $39.06$ & $\bf{23.24}$ & $6$e-$3$ & $33.33$ & $31.40$ & $4$e-$3$ & $33.33$ \\
                  & Init &   & $31.43$ & $8$e-$3$ & $33.33$ & $\bf{23.05}$ & $6$e-$3$ & $33.33$ & $\bf{33.59}$ & $4$e-$3$ & $33.33$ \\
                \multirow{2}{*}{Grad} & Zero &    & $31.78$ & $2$e-$13$ & $25.52$ & $22.74$ & $5$e-$13$ & $25.52$ & $33.25$ & $2$e-$13$ & $35.94$ \\
                  & Init &    & $32.21$ & $7$e-$13$ & $58.33$ & $22.52$ & $4$e-$13$ & $17.71$ & $\bf{33.87}$ & $3$e-$13$ & $42.19$ \\
                \hdashline
                \multirow{2}{*}{Input} & Zero & \multirow{4}{*}{Parameter}  & $\bf{33.12}$ & -- & $30.00$ & $\bf{23.04}$ & -- & $30.00$ & $\bf{34.04}$ & -- & $30.00$ \\
                  & Init &    & $\bf{33.25}$ & -- & $40.00$ & $\bf{23.16}$ & -- & $40.00$ & $\bf{33.96}$ & -- & $60.00$ \\
                \multirow{2}{*}{Grad} & Zero &    & $32.49$ & -- & $10.00$ & $22.19$ & -- & $10.00$ & $32.60$ & -- & $20.00$ \\
                  & Init &    & $32.42$ & -- & $30.00$ & $\bf{22.87}$ & -- & $60.00$ & $32.73$ & -- & $50.00$ \\
            \bottomrule
        \end{tabular}
    \end{adjustbox}
    \caption{Performance difference of Ours$_{\text{ Merge+Del}}$ (XS+WL+TGT) under pruning strategy variations measured on test sets of Japanese Tasks. The best scores (scores with no significant difference from the highest ones) are marked by bold fonts.}
    \label{table:ablation_result}
\end{table*}

\subsection{Training and Inference}
For training on the target tasks, $50$ instances were randomly subsampled for both training and validation sets, respectively, to replicate the low-resource scenario. 
These small subsets were used for training and validating all the methods compared. 
% To assess the impact of varying training data sizes, experiments were conducted with $5$, $100$, and $200$ training instances.
% If either the training or validation data contained fewer than $200$ instances, all available data were used for that split.
LoRA modules for the related tasks were trained using all available training sets. 
The training was stopped early based on the validation loss measured at each epoch. 
The model with the lowest validation loss was saved as the final model. 
Details on LoRA module training parameters are in Appendix~\ref{appendix:implementation_details}.

For testing, all the test set samples were used. 
At inference time, a summary was generated employing greedy decoding.

\subsection{Ablation Study}
We conducted an ablation study to investigate the effectiveness of our design of (a) parameter importance estimation, (b) pruning unit, and (c) pruning value. % and deletion strategies in the Ours$_{\text{ Merge+Del}}$ setting
For (a), we compare our importance calculation method to the one proposed by \citet{zhang2022platon}, which is based on magnitudes of parameter weights and gradients. 
For (b), we compare parameter-wise pruning to module-wise deletion and reinitialization. 
For (c), we examine a method that resets the parameters of pruned modules to their initial values.
Further details on these variations are provided in Appendix~\ref{appendix:pruning}.

%%%%%%%%%%%%%%%%%%%%%%%%%%%%%%%%%%%%%%%%%%%%%%%%%%%%%%%%%%%%
\section{Experiment Results}

Experiments were conducted independently with three different random seeds, and the results are reported as the average across these runs.

\subsection{Main Results}
\label{subsec:main_results}

Table \ref{table:main_result} shows the results of the proposed method and baselines for the $5$ summarization tasks in English and Japanese.\footnote{BERTScore \citep{bertscore} results, which show the consistent tresnds with ROUGE/BLEU scores, are also reported in Appendix~\ref{appendix:additional_results}.}
Remarkably, our method consistently outperforms LoRA and LoRAHub in most tasks across domains and languages.
Comparing Ours$_\text{ Merge}$ and Ours$_\text{ Merge+Del}$, Ours$_\text{ Merge+Del}$ achieves higher performance in $4$ tasks and comparable results in MIMIC-III.
These results clearly confirm the effectiveness of the adaptive LoRA merge that further trains LoRA parameters during merging while pruning ineffective parameters. 
It is noteworthy that the performance gain over LoRAHub is more pronounced on Japanese tasks (Bloomberg, NLP Paper, and Medical Paper), which is another advantage of the proposed method.

% No significant difference is observed between merging only LoRA modules of related tasks and merging both modules of related and target tasks for most datasets.
% Since both approaches use the same small dataset, the benefit of merging target task modules is limited, and repeated training may lead to overfitting.
On Ours$_{\text{ Merge+Del}}$, merging both modules of related and target tasks showed marginal improvements over merging only LoRA modules of related tasks for most datasets. 
We suspect this is because the LoRA modules of related tasks can adapt to the target task through the training during merging. 
The LoRA module of the target task was significantly effective on the Medical Paper dataset, which may imply domain differences matter. 
Further investigation constitutes our future work. 
% We analyze this further in subsection~\ref{subsec:analysis_two_step}.

Table~\ref{table:example_output} shows the generated summaries along with a reference. 
The proposed methods explicitly mention the key innovation, ``\textbf{community-based autoencoders}''.
While Ours$_{\text{ Merge}}$ captures this concept, its description remains vague.
Ours$_{\text{ Merge+Del}}$, however, provides a clearer and more informative summary. 
In contrast, LoRA and LoRAHub generated an overly generalized description of ``\textit{inspired by the way humans learn to communicate},'' which shifts the meaning of ``\textit{Motivated by theories of language and communication.}''
% mainly highlight the inspiration behind the proposed method but 
In addition, they failed to describe the technological novelty, resulting in less sensible summaries for the input paper.

% These results demonstrate that training merged models improves adaptation to the task, and removing ineffective modules further enhances generative quality.

\begin{figure*}[t]
    \centering
    \includegraphics[width=\linewidth]{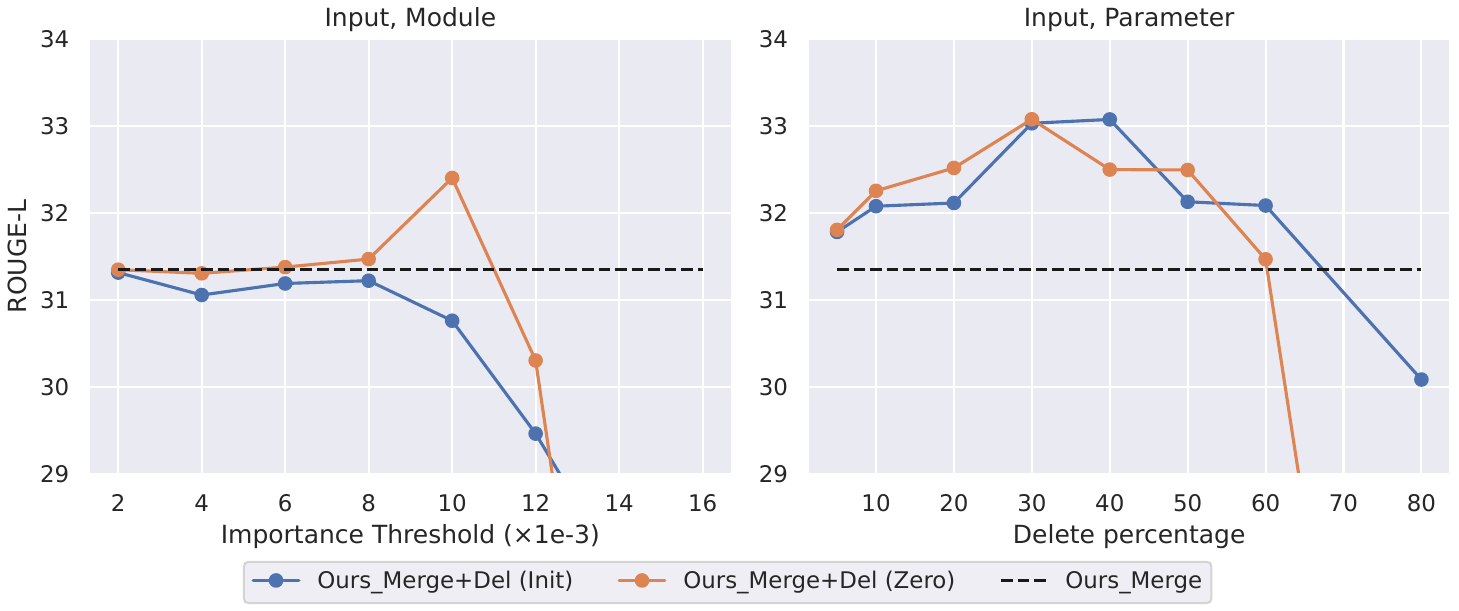}
    \caption{Impact of pruning hyperparameters on model performance (validation set of Bloomberg) %, Ours$_{\text{ Merge+Del}}$ (XS+WL+TGT)).
    }
    \label{fig:ipt_analysis}
\end{figure*}

\subsection{Ablation Study Results}
This section presents the ablation study results on different pruning strategies with the Japanese tasks.
Table~\ref{table:ablation_result} summarizes the model performance measured on the test sets under various pruning configurations: parameter importance calculation method (\textbf{Grad}: magnitudes of parameter weights and gradients; \textbf{Input}: magnitudes of parameter weights and inputs), pruning unit (\textbf{Module}: module-level pruning; \textbf{Parameter}: parameter-level pruning), and pruning values (\textbf{Init}: initialization; \textbf{Zero}: zeroing out).
A baseline without pruning (Ours$_{\text{ Merge}}$ (XS+WL+TGT)) is also included.
The pruning threshold (``Thresh'' column) represents the importance score threshold used for module-level pruning. 
Module-level pruning prunes modules whose average parameter importance score is below the threshold. 
All parameters in a pruned module were reset. 
% For module-level pruning, pruning is performed based on a predefined importance threshold applied uniformly across the entire model, and modules with importance scores below the threshold are reset.
This threshold was treated as a hyperparameter and optimized using validation data.
In contrast, parameter-level pruning prunes $s\%$ parameters of lowest importance scores as shown in the ``Del \%'' column.  

The results indicate that Input, which evaluates parameter importance based on magnitudes of parameter weights and inputs, and Parameter, which conducts parameter-level pruning, consistently achieve higher performance than their counterparts. 
For resetting values on pruning, both methods worked comparably. 
It is noteworthy that pruning with inferior configurations still improved upon the baseline without pruning, which confirms that pruning is crucial in our method.  
% However, the optimal configuration depends on the setting, with no universally superior approach.

To further analyze the effects of pruning configurations, we examine the relationship between pruning hyperparameters and model performance.
Figure~\ref{fig:ipt_analysis} shows the impact of the pruning thresholds on Ours$_{\text{ Merge+Del}}$ (XS+WL+TGT) with module-level (Module) or parameter-level (Parameter) pruning measured on the validation set of Bloomberg. 
The parameter importance was evaluated based on the magnitudes of parameter weights and inputs (Input). 
% the impact of module-level pruning thresholds and parameter-level deletion ratios on the proposed method.
% For module-based pruning, performance fluctuates in the Grad setting, suggesting unreliable importance estimation.
% In contrast, the Input setting shows stable performance at lower thresholds but deteriorates beyond $10$e-$3$, indicating difficulty in distinguishing effective modules.
% However, pruning can still mitigate overfitting by reducing trainable parameters.
The graph of the parameter-level pruning (right) shows a bell-like shape, i.e., the performance initially improves as ineffective parameters are pruned and then decreases when pruning becomes excessive. 
In contrast, the graph of module-level pruning (left) exhibits that the performance hardly outperforms the baseline, which indicates that module-level pruning is too coarse-grained and may result in removing effective parameters in these modules. 
% The Input setting exhibits more stable performance than Grad, implying more reliable importance estimation.
Appendix~\ref{appendix:additional_results} shows the graphs on the Grad configuration. 

\begin{figure}[t]
    \centering
    \includegraphics[width=\linewidth]{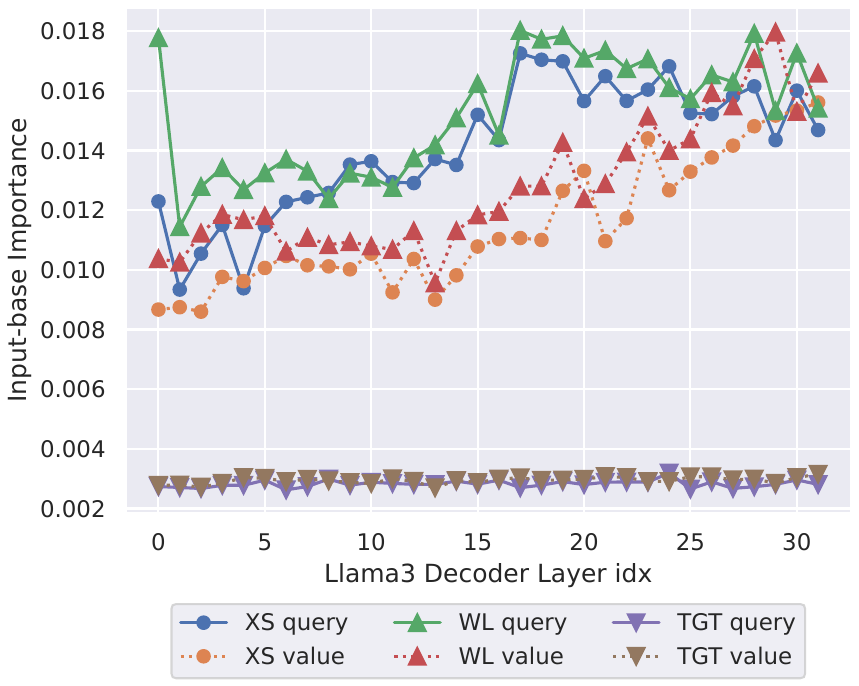}
    \caption{Distribution of Module-wise importance based on Input (Bloomberg, Ours$_\text{ Merge+Del (XS+WL+TGT)}$).}
    \label{fig:module_base_ipt}
\end{figure}

Figure~\ref{fig:module_base_ipt} shows the module-wise importance distribution in different layers of LLM measured on the Bloomberg task, where the importance was calculated based on magnitudes of parameter weights and inputs.
The importance scores of LoRA modules vary: LoRA modules of the target task range from $0.002$ to $0.004$ while those of related tasks range from $0.008$ to $0.018$. 
Also, the score range differs across layers, too. 
This result suggests two things. 
First, for parameter-level pruning, it is crucial to determine pruning parameters per module based on importance score rankings inside a module rather than the global, across-module ranking. 
This aligns with the previous study showing that module-wise importance ranking outperforms global or layer-level pruning in LLM parameter pruning~\citep{sun2024wanda}. 
Second, module-level pruning has a risk of removing target task LoRA modules, which contradicts our expectation that effective parameters should be kept.

\begin{figure}[t]
    \centering
    \includegraphics[width=\linewidth]{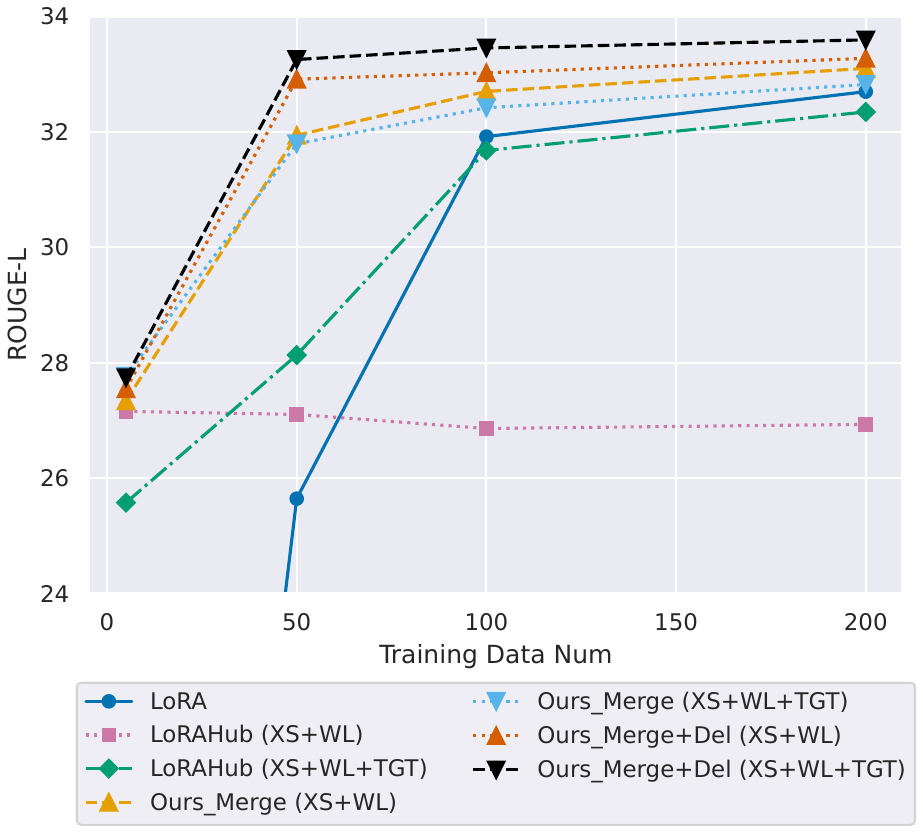}
    \caption{Effect of training data size on model performance (Bloomberg).}
    \label{fig:data_num_analysis}
\end{figure}

\subsection{Effects of Size of Target Task Data}

The previous sections evaluated the performance with a training dataset of $50$ instances on the target task to simulate the low-resource scenario. 
In this section, we investigate the effects of the size of the target training set by varying the size: $5$, $50$, $100$, and $200$ instances on Bloomberg. %and examine its impact on model performance. 
Intuitively, the performance gain by the proposed method should shrink as the training data becomes larger. 

% The experiments were conducted on the Japanese summarization task from Bloomberg.
The results are presented in Figure~\ref{fig:data_num_analysis}. 
As expected, the performance gain by the proposed method shrinks as the training set becomes larger. 
As the number of training instances increases, LoRA (TGT), trained only on the target task, improves significantly.
Yet all the variations of the proposed method still achieve higher ROUGE-L scores across sizes, even at the largest training set, indicating that incorporating LoRA modules from related tasks is useful. 
Furthermore, the proposed method with merging and pruning, Ours$_{\text{ Merge+Del}}$ (XS+WL) and  Ours$_{\text{ Merge+Del}}$ (XS+WL+TGT), consistently outperformed the merging only methods, Ours$_{\text{ Merge}}$ (XS+WL) and Ours$_{\text{ Merge}}$ (XS+WL+TGT), across all data sizes. 
This result again confirms the importance of parameter pruning while merging. 
% These results suggest that while LoRA-based training enables adaptation as data increases, our approach further improves performance by leveraging related tasks effectively.

% \subsection{Case Study}
% To evaluate the effectiveness of our approach, we conduct a case study using the SciTLDR dataset.

%%%%%%%%%%%%%%%%%%%%%%%%%%%%%%%%%%%%%%%%%%%%%%%%%%%%%%%%%%%%
\section{Conclusion}
We proposed the adaptive merging method for multiple LoRA modules to improve LLMs in low-resource tasks.
% Building on prior findings that different LLM layers serve distinct roles, our approach evaluates each LoRA parameter's importance based on weight magnitudes and input activations.
% By pruning less effective parameters and focusing training on the most impactful ones, we enhance the model’s adaptability.
Experiments on the five English and Japanese summarization tasks show that our method significantly outperforms existing LoRA merging techniques across domains and languages.
% The results confirm that adaptive module selection and integration improve task performance.

Future work includes the application of the proposed method to broader tasks and cross-lingual settings. 
Additionally, we plan to evaluate its effectiveness across various LLMs of different sizes.
Exploring the merging of more diverse and numerous LoRA modules is another important direction. 
Currently, the proposed method requires tuning the pruning threshold for each task. 
Automating this process would enhance the practicality of our method.

\section*{Limitations}
Our method conducts LoRA training twice: once to pre-train them for related tasks and another to merge, leading to increased training time.
Although the merging step on the target task is efficient, as we assume the low-resource scenario (in our experiments, we used just $50$ instances), the overall cost remains a concern.
This could be mitigated by leveraging publicly available pre-trained LoRA adapters.

We experimented with summarization tasks in English and Japanese, but summarization itself was monolingual. 
It is worth investigating the applicability of the proposed method to cross-lingual tasks. 

Another limitation is that the proposed method requires tuning the hyperparameter of the pruning ratio, which should be adjusted depending on the datasets. 
Future work should explore automatic methods to determine this hyperparameter. 
% Lastly, we did not analyze the detailed training dynamics of our two-stage approach.
% Understanding its impact could provide insights into LLM learning mechanisms, which we leave for future research.

\section*{Acknowledgments}
We sincerely appreciate the anonymous reviewers for their insightful comments and suggestions to improve the paper. 
This work was supported by the Cross-ministerial Strategic Innovation Promotion Program (SIP) on “Integrated Health Care System” Grant Number JPJ012425.

\bibliography{custom}

\appendix
\section{Details of Dataset Construction}
\label{appendix:dataset_construction}
This section details the construction processes for the Bloomberg, NLP/Medical Paper, and MIMIC-III datasets.

\subsection{Bloomberg Dataset}
The Bloomberg dataset was constructed from Japanese news articles published on Bloomberg's online platform.
The primary goal was to create a task structurally distinctive from the XLSum task by utilizing article highlights as summaries instead of lead sentences.
The dataset was constructed through the following steps:

\begin{enumerate}
    \item \textbf{Article Collection}:  
    We referred to the URL list provided by the MassiveSumm project \cite{varab2021massivesumm}, which includes links to Bloomberg articles.
    Articles containing bullet-point highlights were identified and extracted for further processing.

    \item \textbf{Highlight Extraction}:  
    The bullet-point highlights, a key feature of Bloomberg articles, were automatically extracted using an HTML parser.
    These highlights summarize the essential points of the article and were used as the basis for the output summaries.

    \item \textbf{Title Combination}:  
    To enhance coverage, the extracted highlights were combined with the article title.
    This combination ensures that the summary captures the main content more comprehensively, as the highlights alone may sometimes lack sufficient detail.

    \item \textbf{Input Document Construction}:  
    The full text of each article was extracted and used as the input document.
    This includes all relevant content except for metadata or sections not related to the main article text.
\end{enumerate}

This construction approach differs from that used in other datasets, such as MassiveSumm and XLSum.
While MassiveSumm extracts summaries from lead sentences, they may contain extraneous information not found in the main article. 
Our method leverages bullet-point highlights that are closely tied to the core content.
This ensures a more accurate representation of the article and introduces structural variety between the target and related tasks.

\subsection{NLP/Medical Paper Dataset}
We constructed two datasets for research paper summarization: one using NLP research papers and the other using medical case reports.

\subsubsection{NLP Paper Dataset}
The NLP Paper dataset was created from the LaTeX corpus of the Journal of Natural Language Processing.
The construction process involved the following steps:

\begin{enumerate}
    \item \textbf{Document Extraction}:  
    We extracted LaTeX source files from the corpus, selecting only papers written in Japanese.

    \item \textbf{Title and Abstract Extraction}:  
    The title was extracted from either the `\texttt{jtitle}' or `\texttt{title}' field, while the abstract was extracted from either the `\texttt{jabstract}' or `\texttt{abstract}' field.

    \item \textbf{Preprocessing}:  
    LaTeX-specific commands such as `\texttt{\textbackslash cite}' and `\texttt{\textbackslash vspace}' were removed. %to ensure compatibility with the title generation task.
\end{enumerate}

\subsubsection{Medical Paper Dataset}
The Medical Paper dataset was constructed from case reports published on J-STAGE.
The dataset construction involved:

\begin{enumerate}
    \item \textbf{Document Collection}:  
    Case reports from multiple journals were collected to cover diverse topics.

    \item \textbf{Title and Abstract Extraction}:  
    Titles and abstracts were extracted automatically from the structured metadata of each report.
\end{enumerate}

\begin{table}[t]
    \centering
    \begin{adjustbox}{max width=\linewidth}
        \begin{tabular}{lr}
            \toprule
                Parameter & Value \\
            \midrule
                LoRA Rank & $8$ \\
                LoRA Alpha & $32$ \\
                LoRA Dropout & $0.05$ \\
                Target Modules & Query, Value \\
            \midrule
                Learning Rate & $0.0001$ \\
                Optimizer & AdamW \\
                Batch Size & $16$ \\
                Epoch Num & $40$ \\
            \bottomrule
        \end{tabular}
    \end{adjustbox}
    \caption{Parameters used for LoRA module training.}
    \label{table:training_params}
\end{table}

\subsection{MIMIC-III Dataset Processing}
For the MIMIC-III dataset, we extracted and processed radiology reports for the summarization task following the methodology proposed in RadAdapt \citep{van2023radadapt}.
The procedure consisted of the following steps:

\begin{enumerate}
    \item \textbf{Section Extraction}:  
    We extracted the \textit{Findings} and \textit{Impressions} sections from raw radiology reports.
    The \textit{Findings} section serves as the input, while the \textit{Impressions} section, which provides a concise summary of key observations, serves as the output.

    \item \textbf{Filtering}:  
    To further refine the dataset, we applied an additional filtering step.
    Specifically, samples where the \textit{Findings} section was shorter than or comparable in length to the \textit{Impressions} section were removed, ensuring that the dataset aligns with the characteristics of a summarization task.
\end{enumerate}

This filtering step improves dataset quality by ensuring that the input text contains more detailed information than the output summary, reinforcing a meaningful document-summarization relationship.

\begin{table*}[t]
    \centering
    \begin{adjustbox}{max width=\linewidth}
        \begin{tabular}{lp{0.7\linewidth}}
            \toprule
                Dataset & Prompt \\
            \midrule
                XLSum & Summarize the following Article in no more than three sentence. \\
                        & Article: \{\{article\}\} \\ 
                        & Summary: \\
                WikiLingua & Summarize the following How-to Guide and write a one-sentence summary for each step: \\ 
                        & How-to Guide: \{\{article\}\} \\
                        & Summary: \\
            \hdashline
                Bloomberg & Summarize the following article in three sentences. \\
                        & Article: \{\{article\}\} \\ 
                        & Summary: \\
                Title Generation & Read the following Abstract of a scientific paper and create an appropriate title that reflects the content. Please only output the Japanese title. \\ 
                        & Abstract: \{\{article\}\} \\ 
                        & Title: \\
                MIMIC-III & Summarize the following radiology report. \\
                        & Findings: \{\{article\}\} \\
                        & Impression: \\
                SciTLDR & Write a TLDR by summarizing the following scientific paper in one sentence based on its Key Sections (Abstract, Introduction, and Conclusion). \\
                        & Key Sections: \{\{article\}\} \\
                        & TLDR: \\
            \bottomrule
        \end{tabular}
    \end{adjustbox}
    \caption{Prompt Design}
    \label{table:prompt}
\end{table*}

\section{Implementation Details}
\label{appendix:implementation_details}
\subsection{LoRA Training Parameters}
Table~\ref{table:training_params} presents the parameters used for LoRA module training.

\subsection{Computation Environment}
Experiments were conducted on NVIDIA RTX A$6000$ GPUs with $48$GB of memory.
We used $2$ GPUs for training LoRA modules and merging them under the proposed method, while $1$ GPU was allocated for training baseline methods such as LoRAHub and for inference.

\subsection{Prompt Design}
Table~\ref{table:prompt} presents the prompt design used in both LoRA training and output generation.

\begin{table*}[t!]
\centering
%\resizebox{\textwidth}{!}{%
\begin{tabular}{@{}lccccc@{}}
\toprule
                            & \multicolumn{1}{l}{MIMIC-III} & \multicolumn{1}{l}{SciTLDR} & \multicolumn{1}{l}{Bloomberg} & \multicolumn{1}{l}{NLP Paper} & \multicolumn{1}{l}{Medical Paper} \\ \midrule
Zero-shot                   & $0.693$                         & $0.739$                       & $0.605$                         & $0.627$                         & $0.637$                             \\
LoRA (XS)                   & $0.729$                         & $0.601$                       & $0.692$                         & $0.754$                         & $0.776$                             \\
LoRA (WL)                   & $0.698$                         & $0.756$                       & $0.717$                         & $0.797$                         & $0.812$                             \\
LoRA (TGT)                  & $0.763$                         & $0.778$                       & $0.710$                         & $0.817$                         & $0.843$                             \\
LoRAHub(XS+WL)              & $0.717$                         & $0.745$                       & $0.719$                         & $0.798$                         & $0.809$                             \\
LoRAHub (XS+WL+TGT)         & $0.763$                         & $0.780$                       & $0.726$                         & $0.824$                         & $0.827$                             \\
Ours$_{\text{ Merge}}$ (XS+WL)         & $0.768$                         & $0.782$                       & $0.750$                         & $0.824$                         & $0.840$                             \\
Ours$_{\text{ Merge}}$ (XS+WL+TGT)     & $0.769$                         & $0.780$                       & $0.749$                         & $0.820$                         & $0.843$                             \\
Ours$_{\text{ Merge+Del}}$ (XS+WL)     & $0.766$                         & $0.783$                       & $0.752$                         & $0.838$                         & $0.840$                             \\
Ours$_{\text{ Merge+Del}}$ (XS+WL+TGT) & $0.766$                         & $0.783$                       & $0.757$                         & $0.825$                         & $0.857$                             \\ \bottomrule
\end{tabular}%
%}
\caption{BERTScore results on five summarization tasks of various domains and multiple languages.}
\label{tab:bertscore}
\end{table*}

\begin{figure*}[t!]
    \centering
    \includegraphics[width=\linewidth]{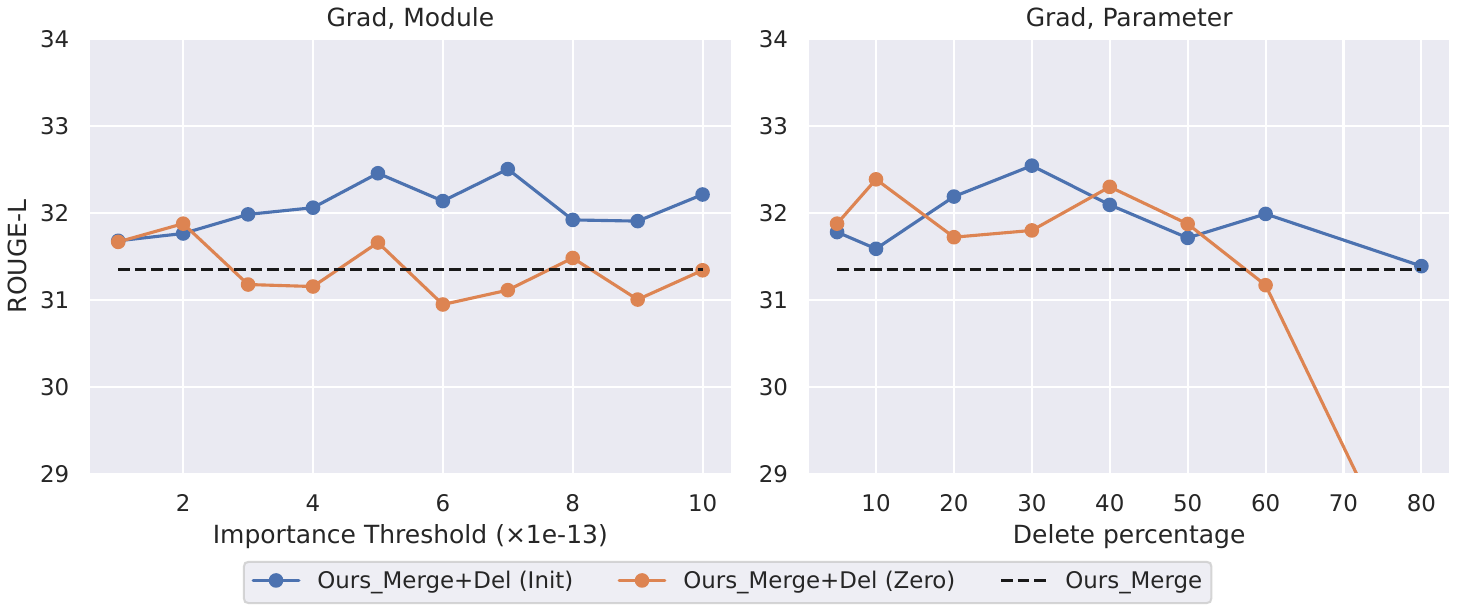}
    \caption{Impact of pruning hyperparameters on model performance (Bloomberg, Ours$_{\text{ Merge+Del}}$ (XS+WL+TGT), Grad).}
    \label{fig:ipt_analysis_appendix}
\end{figure*}

\section{Pruning Strategies}
\label{appendix:pruning}
% This section describes the additional pruning strategies explored in the Ours$_{\text{ Merge+Del}}$ setting.

% \subsection{Calculation of Parameter Importance}
As the proposed method, we used the importance evaluation metric based on magnitudes of parameter weights and inputs.
In the ablation study, we compared it to another metric that considers the magnitudes of parameter weights and gradients.
% To ensure compactness in the appendix, we remove explicit notation of $W_{ij}$ from the equations while still using it for clarity in expressions.
This metric is defined as follows:
\begin{align*}
    I &= |W_{ij} \cdot \Delta W_{ij}|
\end{align*}
where $\Delta W_{ij}$ represents the gradient of weight $W_{ij}$.
This formulation estimates the impact of pruning $W_{ij}$ by approximating the change in loss when setting $W_{ij}$ to zero \citep{molchanov2019importance, liang2021supertickets}.

To address the variance caused by batch sampling, we apply an uncertainty-aware smoothing technique \citep{zhang2022platon, zhang2023adalora}.
The importance at step $t$, denoted as $I^{(t)}$, is smoothed using an exponential moving average to obtain $\bar{I}^{(t)}$.
Additionally, the uncertainty measure $\bar{U}^{(t)}$ quantifies the local fluctuations of $I^{(t)}$.
The final importance score $S^{(t)}$ is computed as the product of these two terms:
\begin{align*}
    \bar{I}^{(t)} &= \beta_1 \bar{I}^{(t-1)} + (1-\beta_1)I^{(t)} \\
    \bar{U}^{(t)} &= \beta_2\bar{U}^{(t-1)} + (1-\beta_2) | I^{(t)} - \bar{I}^{(t)} | \\
    S^{(t)} &= \bar{I}^{(t)} \cdot \bar{U}^{(t)}
\end{align*}

% \subsection{Pruning Method}
% In the methodology, we proposed zeroing-based pruning to remove the influence of low-importance LoRA modules.
% As an additional strategy, we introduce reinitialization-based pruning, where the parameters of pruned modules are reset to their initial values before training.
% This method mitigates abrupt parameter changes that can destabilize the model and allows previously pruned modules to potentially regain importance during later stages of training.

% \subsection{Pruning Unit}
% In the methodology, we proposed parameter-wise pruning, where individual weights within a LoRA module are selectively removed.
% As an alternative, we introduce module-wise pruning, where the entire LoRA module is evaluated and removed if its importance falls below a predefined threshold.
% The importance of a module is determined by aggregating the importance of its weights, and a global threshold is set across all modules.
% This approach simplifies pruning decisions and reduces computational overhead while effectively eliminating redundant LoRA modules.

\section{Additional Results}
\label{appendix:additional_results}
Table~\ref{tab:bertscore} shows BERTScore results. 
Figure~\ref{fig:ipt_analysis_appendix} shows the impact of the pruning thresholds on Ours$_{\text{ Merge+Del}}$ (XS+WL+TGT) with Grad and Module or Parameter level pruning configurations.

\end{document}